\documentclass{article}
\usepackage{ijcai20}
\usepackage{comment}

\usepackage{times}
\usepackage{soul}
\usepackage{url}
\usepackage[utf8]{inputenc}
\usepackage[small]{caption}
\usepackage{graphicx}
\usepackage{amsthm}
\usepackage{booktabs}
\usepackage{algorithm}
\usepackage{algorithmic}
\usepackage{mathrsfs}
\usepackage{mathtools}
\usepackage{xcolor}
\usepackage{breqn}
\usepackage{amssymb}
\usepackage{amsmath}

\title{Attribute2vec: Deep Network Embedding Through Multi-Filtering GCN}

\author{
Tingyi Wanyan$^1$
\and
Chenwei Zhang$^1$\and
Ariful Azad$^1$\and
Xiaomin Liang$^3$\and
Daifeng Li $^3$\and
Ying Ding$^{1,2}$
\affiliations
$^1$Indiana University Bloomington\\
$^2$University of Texas at Austin\\
$^3$Sun Yat-Sen University, Guangzhou, China\\
\emails
\{tiwanyan, zhang334, azad\}@iu.edu,
ying.ding@ischool.utexas.edu,
liangxm23@mail2.sysu.edu.cn,
lidaifeng@mail.sysu.edu.cn
}

\begin{document}
\maketitle
\begin{abstract}
We present a multi-filtering Graph Convolution Neural Network (GCN) framework for network embedding task. It uses multiple local GCN filters to do feature extraction in every propagation layer. We show this approach could capture different important aspects of node features against the existing attribute embedding based method. We also show that with multi-filtering GCN approach, we can achieve significant improvement against baseline methods when training data is limited. We also perform many empirical experiments and demonstrate the benefit of using multiple filters against single filter as well as most current existing network embedding methods for both the link prediction and node classification tasks. 
  
\end{abstract}

\section{Introduction}
Network embedding aims to embed network information into low-dimensional vector space so that standard machine learning algorithms can be applied directly to investigate latent features of networks. To ensure the quality of various network mining tasks, the embedded vector representation should be representative--it needs to encode as much information about a node or an edge as possible, including its context information as well as the structure information. Such a vector should also be concise--the dimension of it should not be too large, in consideration of both memory usage efficiency and computational efficiency. Getting appropriate embedding becomes more challenging nowadays since the network size and the node attribute variety both grow exponentially fast. It is crucial to gain a representation to embed diverse node attributes as well as the structural properties from a network.

Among different network embedding methods, pioneer works such as DeepWalk \cite{perozzi2014deepwalk}, and node2vec \cite{grover2016node2vec} have laid a foundation for embedding framework that incorporates deep learning and skip-gram architecture \cite{mikolov2013efficient}. Such works propose to use random walk for simulating the context of a center node in the network. Following this architecture, subsequent techniques, such as Metapath2vec \cite{dong2017metapath2vec} also solves the heterogeneous network embedding problem. In most of these methods, various types of nodes are used to generate random walks and incorporated into the objective function. Methods of this kind have flexible strategies of generating different random walks, and perform well in capturing neighborhood similarities and detecting community membership. However, most of them only focus on encoding network structure, which take the whole network structure as input to generate embedding vectors, while keeping little attention on node attributes that also provide important information. Most of these techniques focus on transductive embedding, which typically operates on fixed graph thus does not generalize to unseen data.

Node attributes in network embedding have been proven to be important \cite{huang2017accelerated}; however, the embedding of networks consisting of nodes affiliated with various attributes is still in its early stage due to the complexity of attributes \cite{zhang2018anrl}. There are a few works combining the network structure and node attributes to build their embeddings, such as 
LINE \cite{tang2015line} and AANE \cite{huang2017accelerated}. 
Different methods have been adopted to extract the node attributes information. Most of these models are limited in their transductive essence--the input to the system has to be the whole fixed network with node attributes, which causes the model to generate poorly from the training data to unseen data. Another limitation is the scalability--such methods need to perform computation with respect to matrix that scales linearly to the graph size.

To address the aforementioned issues, recently proposed GraphSAGE model \cite{hamilton2017inductive} uses local aggregation functions for embedding computation, which largely simplifies the whole graph propagation procedure mentioned above, and also achieves inductive property. Inspired by this framework, we build \textit{MF-GCN}, which adopts a multiple-filtering local GCN model as the aggregation function, incorporates a skip-gram with negative sampling model \cite{levy2014neural}, and a quadratic mean-square-error on the node attributes as the objective function. We demonstrate that with multiple-filtering GCN architecture, the system can capture diverse aspects of node attributes. Two network learning tasks, node classification and link prediction, are conducted in four benchmark datasets to evaluate the performances of our algorithm, compared to several other state-of-the-art approaches. The contributions of this research is summarized as below:

\begin{enumerate}
    \item We introduce a novel approach called  \textit{MF-GCN} based on multi-filter GCN aggregator for extracting different aspects of node attributes. \textit{MF-GCN} outperforms state-of-the-art methods in capturing diverse node attributes, along with the network structure information.
    \item We show that \textit{MF-GCN} can achieve significantly better performance than other baseline methods when the training data is limited.
    \item We empirically demonstrate the impact of using different numbers of GCN filters and suggest optimal numbers of filters for different learning tasks.
    \item We conduct various experiments and show that our \textit{MF-GCN} model has superior performance over most of the baseline methods on link prediction and node classification tasks.
\end{enumerate}


\section{Literature Review}
Representation learning on network data has mainly been about performing embedding operation, which involves an \textbf{Encoding} and a \textbf{Decoding} parts. Encoding generates an embedding, and decoding is used as the objective function to update model learning parameters. Techniques to solve this problem nowadays usually follow three tracks: \textbf{Skip-gram-based model}; \textbf{Matrix Factorization model}; and \textbf{Graph Neural Network}. Below we review related research in these three tracks.


\textbf{Skip-gram-based model} defines a framework for embedding by incorporating the information of relevant nodes appearing in certain steps of random walk of the target node. 
Many existing methods have adopted such model. For example, DeepWalk \cite{perozzi2014deepwalk} adopts uniform sampling strategy for performing random walk, then uses the skip-gram model with negative sampling for computing the objective function. However, the uniform sampling strategy and unclearly defined objective function could not well preserve the network structure. Node2vec \cite{grover2016node2vec} combines Breadth-first and Depth-first searching strategies, so it remains more first- and second-order proximity; but still it lacks a clear objective function for preserving global network structure. LINE \cite{tang2015line} defines an objective function to minimize the KL-divergence between the empirical distribution and the objective distribution; thus, it does well on preserving first- and second-order proximity, but the local sampling strategy still lacks information for global network structure. HARP \cite{chen2018harp} 
shows the improvement of using a graph coarsening strategy on different skip-gram approaches.

\textbf{Matrix Factorization method} performs embedding through factorizing a similarity matrix, and uses the factorized vector as the embedding representation vector. This stream of research has mostly focused on designing a similarity matrix to perform factorization. Typically the operation is eigen-decomposition or singular-value decomposition. Representative work in this area includes 
 HOPE \cite{ou2016asymmetric} and Netsmf \cite{qiu2019netsmf}.

\textbf{Graph Neural Network} defines a framework through graph convolution to incorporate neighborhood information into the encoding layer. From the approach perspective, this line of work could be categorized into two branches: Spectral-based Filtering and Spatial-based Filtering. Spectral-based Filtering method performs graph convolution and forms the proposition function in the spectral domain. ChebNet \cite{defferrard2016convolutional} follows Spectral Graph Theory \cite{hammond2011wavelets} to define a spectral graph convolution with the full set of Chebyshev Polynomial approximation to the convolution kernel. Graph Convolutional Neural Network (GCN) \cite{kipf2016semi} later simplifies it by using only first-order approximation. FastGCN \cite{chen2018fastgcn} uses a sampling technique to avoid whole graph convolution. Following the GCN idea, AGCN \cite{li2018adaptive} adopts its structure but creates different learning metrics to approximate a Gaussian kernel. Spatial-based Filtering method performs graph convolution in a spatial domain. The original work \cite{scarselli2008graph} defines a framework to recursively update propositional function until an equilibrium point is reached. GGNN \cite{lin2015learning} improves this model by conducting a gated, recurrent unit. Later various techniques have been proposed in different ways, including designing efficient sampling strategy \cite{chen2018fastgcn}, improving the model inductive ability \cite{hamilton2017representation}, and defining convolution operation on spatial domain \cite{niepert2016learning}. Graph Attention Mechanism \cite{velivckovic2017graph} later comes to further improve the model by assigning attention coefficients to address relative importance of neighborhood nodes.
\section{Preliminary Requirement and Optimization Model}
In this section, we present the prerequisite models and techniques for our method. We first define the problem, then review the Graph Convolutional Neural Network model and its local extended version that is crucial to our approach. Later the optimization model used in our method to update the system parameters is derived.
\subsection{Problem Definition}
Generally, a network is defined as $G=\{V,X,W,E\}$, where $V$ is a finite set of vertices ($|V|=N$, $N$ the number of nodes); $E$ is the set of edges; $X$ is the set of attributes associated with each node; $W$ is the weighted adjacency matrix, where each entry $W_{ij}$ represents the weight of each edge. In this work, we focus on the unweighted homogeneous network embedding, so $W_{ij}=1$ if there is an edge $e=(i,j)$ connecting vertices $i$ and $j$, otherwise $W_{ij}=0$. we define $f:V\rightarrow{R^{d}}$ as an embedding function that maps each node to a latent representation. For every center node $u\in V$, let $N_c(u)\subset V$ denote its context nodes generated from a random walk sampling strategy, and $N(u)\subset V$ denote the one-hop neighborhood nodes of center node $u$. The problem we try to solve is to find the best $f$ to generate embedding vectors for nodes which could fulfill various machine learning tasks. 

\subsection{GCN and Its Local Version}
In GCN, the graph convolution operation is used as layer propagation function,
which could be written as:
\begin{equation}
    Z=D^{-\frac{1}{2}}AD^{-\frac{1}{2}}X\theta
    \label{GCN_eq}
\end{equation}
where $Z\in R^{N\times F}$ is the convolved embedding feature matrix (the intermediate layer hidden representation), $A$ the graph adjacency matrix plus an Identity matrix, and $D$ the diagonal degree matrix where $D_{ii}=\sum_{j}A_{ij}$, and $X\in R^{N\times C}$ the signal with C input channels (C-dimensional attribute vector for each node). $\theta \in R^{C\times F}$ is the learnable matrix of filter parameters, $N$ the total number of nodes in the graph, and $F$ the dimension of the embedding vectors. 

\textbf{Local GCN} \cite{hamilton2017inductive} is an extended local version of GCN.
Since $A$ in Equation \ref{GCN_eq} is the adjacency matrix and $D$ is the diagonal degree matrix, this formula could actually be treated as a normalization function that each entry of the output vector represents a node status propagated by all of its neighbor nodes. Equation \ref{GCN_GNN} represents this operation:
\begin{equation}
    h^{l+1}_{i}=\sigma(\sum_{j\in N(i)}\frac{1}{\sqrt{D(i,i)D(j,j)}}h^{l}_{j}\theta^{l})
    \label{GCN_GNN}
\end{equation}
where $h^{l}_{i}\in R^{F}$ is the hidden representation of node $v_i$ in $l^{th}$ layer, $\theta^{l}\in R^{C\times F}$ is the weight parameters in $l^{th}$ layer, and $F$ is the dimension of the embedding vector. $N(i)$ is the neighborhood nodes of a center node $v_i$. This is a per-neighborhood normalization of the parameterized propagation layer. In the current work, we extend to use multiple local GCN filters as the propagation function.

\subsection{Optimization Model}
We follow the transductive optimization model from \cite{yang2016revisiting}, aiming at preserving the network structure information as well as predicting the correct node label. 
To preserve structure information, we apply the skip-gram model, which seeks to maximize the probability of observing the context nodes $N_c(u)$ of a center node $u$ based on its embedding representation:   
\begin{equation}
    \mathrm{max}_{f}\sum_{u\in V}Pr(N_c(u)|f(u))
    \label{obj}
\end{equation}
We follow the assumption of conditional independency \cite{grover2016node2vec}, where the probability of observing one context node is independent of other context nodes given the center node:
\begin{equation}
    Pr(N_c(u)|f(u))=\prod_{c_i\in N_{c}(u)}Pr(c_{i}|f(u))
\end{equation}
where $c_{i}$ is the $i\mathrm{th}$ context node of the center node. 
We maximize the objective function (Equation \ref{obj}) by minimizing its negative log form:
\begin{equation}
    \mathcal{L}_{s}=-\sum_{u\in V}\sum_{c_{i}\in N_{c}(u)}\mathrm{log}Pr(c_{i}|f(u))
    \label{loss_NCE}
\end{equation}
where 
$Pr(c_{i}|f(u))$ is the prediction probability. Let $\vec{c_{i}}$, $\vec{u}$ denote the embedding vector representation of context node $c_{i}$ and the associated center node $u$ that are mapped by $f$, we model this probability as follows:

\begin{equation}
        Pr(c_{i}|f(u))=\frac{e^{\vec{c}_{i}\cdot \vec{u}}}{Z_{u}}
    \label{soft_max}
\end{equation}
    
where $Z_{u}=\sum_{c_j\in V}e^{\vec{c_{j}}\cdot \vec{u}}$ is the normalization factor that integrates over all nodes. $\vec{c}_{i}\cdot \vec{u}$ is the inner product between $\vec{c}_{i}$ and $\vec{u}$ that represents the similarity of these two embedding representations. With this assumption, Equation \ref{loss_NCE} could be simplified to:
\begin{equation}
    \mathcal{L}_{s}=-\sum_{u\in V}\Big[\sum_{c_{i}\in N_{c}(u)}\vec{c_{i}}\cdot \vec{u}-logZ_{u}\Big]
    \label{sim_loss}
\end{equation}
Numerical computation of $Z_{u}$ is huge for large graph with millions of nodes, since the computation grows linearly with graph size. So we adopt negative sampling to approximate the normalization factor; thus the objective function becomes:

\begin{equation}
\label{SGNN_loss}
\begin{split}
     \mathcal{L}_{s}&=-\sum_{u\in V}\Big[\sum_{c_{i}\in N_{c}(u)}log\sigma(\vec{c_{i}}\cdot \vec{u})\\
     &+\sum_{j=1}^{\mathbb{K}}E_{c_{j}\sim P_{v}(c_{j})}log\sigma(-\vec{c_{j}}\cdot \vec{u})\Big]
\end{split}
\end{equation}

where $\sigma(x)=\frac{1}{1+\exp(-x)}$, $\mathbb{K}$ is the number of negative samples. $P_{v}(c_{j})$ is the negative sampling distribution.

For supervised loss of predicting the correct label, we adopt a cross entropy loss on a softmax layer:
\begin{equation}
    \mathcal{L}_r=-\sum_{i=1}^{M}y_{i}log(p_{i}(u))
\end{equation}
where $M$ is the class number, $y_{i}$ is the ground truth node label, and $p_{i}(u)$ is the predicted label of center node $u$ from softmax layer.

\section{Methodology}
In this section, we present the framework of our \textit{MF-GCN}, and the whole system architecture. We achieve this by first filtering node attributes with multiple local GCN filters, then concatenating these filtered hidden representation vector to form one hidden layer; the succeeding hidden layer follows the same procedure. 

\subsection{Multiple-Filtering GCN Aggregator}



\begin{figure}[t]
        \begin{center}
            \includegraphics[scale=0.6]{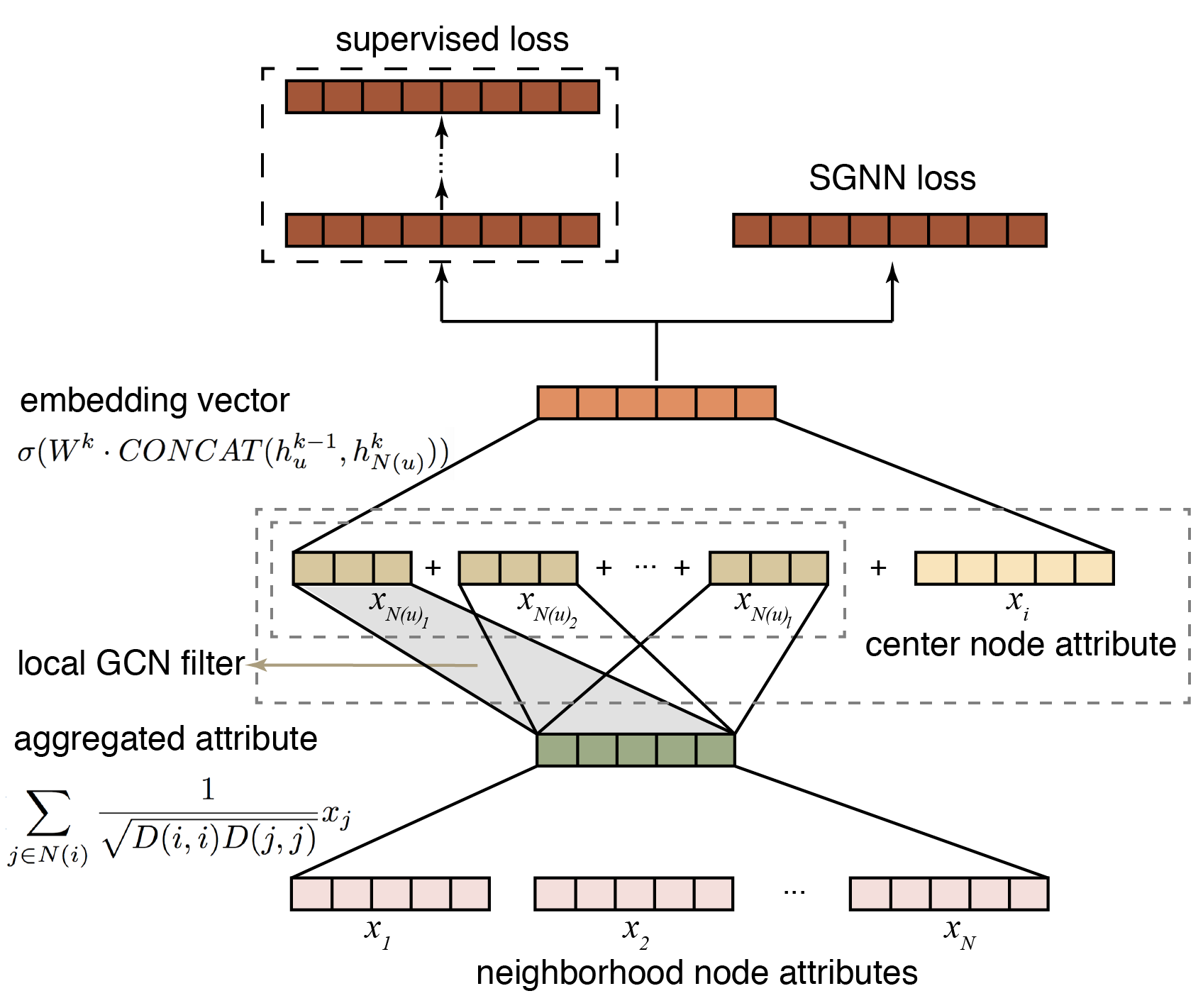}
        \end{center}
        \caption{\textit{MF-GCN} model architecture. ('+' means concatenation).}
        \label{MF_gcn}
\end{figure}

Since local GCN could propagate information from layers to layers, it is intuitive to assume that multiple such local GCN filters could propagate information from different aspect of node attributes. 
Therefore we propose a multi-filtering GCN architecture--each single local GCN filter is one attribute extractor, so multiple such filters form one layer of feature extraction. Figure \ref{MF_gcn} visually shows this operation. For each hidden layer, we use multiple distinct local GCN filters (each filter has distinct parameters) to operate on the input of the previous hidden layer; for the first layer, the operation is performed on the input from the node attribute channels. This operation is shown below:


\begin{equation}
    x_{N(i)_{l}}=\alpha(\sum_{j\in N(i)}\frac{1}{\sqrt{D(i,i)D(j,j)}}x_{j}\theta_{l})
    \label{GCN_aggregate_pre}
\end{equation}
where $x_{i}\in R^{F}$ is the input attribute channel vector for node $i$, $x_{N(i)_l}$ is the aggregation output from the $l$-th local GCN aggregator, 
$\theta_{l}$ is the learnable parameters for the $l$-th local GCN aggregation function, and $\alpha()$ is the non-linear activation function. 
After concatenating these aggregated representation vectors from different local GCNs to form the final aggregated feature vector, we define our \textit{MF-GCN} aggregation function as below:
\begin{equation}
    h^{k}_{N(u)}=CONCAT(x_{N(u)_{1}},x_{N(u)_{2}},...,x_{N(u)_{L}}))
\end{equation}
where $h^{k}_{N(u)}$ represents the aggregation output for node $u$ in the $k$-th layer. $L$ is the total number of Local GCN Filters. Following the forward propagation algorithm in \cite{hamilton2017inductive}, the final generated embedding vector is derived as:
\begin{equation}
    h^{k}_{u}=\sigma(W^{k}\cdot CONCAT(h^{k-1}_{u},h^{k}_{N(u)}))
\end{equation}
Where $h^{k}_{u}$ represents the hidden embedding representation in the $k-th$ layer, $W^{k}$ is the parameter of the fully connected encoding layer, and $\sigma$ is the non-linear activation function.

\subsection{Network Embedding Framework}
We design our network embedding system according to the Graph Auto Encoding framework. Algorithm \ref{alg:MG_GCN} shows the whole procedure of the \textit{MF-GCN} embedding generation algorithm we propose. Figure \ref{MF_gcn} visually shows the whole system structure. The system can be divided into the encoding and decoding parts.

\textbf{Encoding}: In the encoding phase, the input is the node attribute vector $x_u$. The \textit{MF-GCN} embedding generation procedure is performed on the input vector as shown in Algorithm \ref{alg:MG_GCN}, and finally get the embedding vector $z_u$

\textbf{Decoding}: The decoding phase includes two parts, a supervised loss $\mathcal{L}_r$ and a Skip-gram model with Negative Sampling error (SGNN) $\mathcal{L}_s$. These two loss function are integrated to form the final loss function:
\begin{equation}
    \mathcal{L}=\mathcal{L}_s+\mathcal{L}_r
\end{equation}
The supervised loss part is constructed by a fully connected layer followed by softmax layer for classifying different node labels, it is separated from the SGNN loss part. The input of both parts comes from the embedding output representation $z_u$.

\subsection{Training procedure}
The mini-batch gradient descent is applied in the training process. For each training batch, we randomly select batch sized number of nodes as the training center nodes. Random walk is performed to sample the context nodes for each center node $u$. Here we use the random walk strategy according to node2vec, and the random walk length is set to be $20$. For negative sampling part, we uniformly sample nodes that do not have connections with those one-hop neighbor nodes of the center node; neither do they have connections with the context nodes sampled previously. The negative sample number is set to be 100 for each center node.

Training is performed jointly between the two loss functions $\mathcal{L}_s$ and $\mathcal{L}_r$. For the supervised loss $\mathcal{L}_r$, the training is straightforward by incorporating a softmax layer with cross-entropy loss. For the SGNN loss function part, at each training epoch, we first propagate the center node and derive its embedding representation $\vec{u}$. After sampling its context nodes, we input all these nodes to the system and derive the embedding representations of all context nodes $\vec{c_{i}},c_{i}\in N_{c}(u)$. Then we perform negative sampling and input these negative samples into the system to get the embedding representations for these negative sampled nodes $\vec{c_{j}},c_{j}\in N_{neg}(u)$. Then we calculate the SGNN loss function according to \ref{SGNN_loss}. Note that we use all the normalized form of these embedding vector to perform inner product, since we are measuring their cosine similarities.
\begin{algorithm}[tbh]
\caption{\textit{MF-GCN} embedding generation}
\label{alg:MG_GCN}
\textbf{Input}: Graph $G(V,X,E)$; input node attributes $\{X_{u},\forall{u}\in V\}$; network layer depth K; number of local GCN filters $L$ in each layer; initialized weight matrices $W$; non-linear activation function $\alpha,\sigma$.\\
\textbf{Output}: Embedding vector representation for all $u\in V$
\begin{algorithmic}[1] 
\STATE $h^{0}_{u}=X_{u},\forall{u}\in V$.
\WHILE{not converged}
\STATE sample a mini-batch of center nodes $u$.
\STATE sample the corresponding context nodes $N_{c}(u)$ of these center nodes through certain random walk strategy.
\STATE perform negative sampling on these center nodes according to negative sampling distribution $P_{v}(c_{j})$.
\FOR{k=1...K}
\STATE $h_{N(u)_{l}}=\alpha(\sum_{j\in N(u)}\frac{1}{\sqrt{D(u,u)D(j,j)}}h_{j}^{k}\theta_{l}),l\in L$
\STATE $h^{k}_{N(u)}=CONCAT(h_{N(u)_{1}},h_{N(u)_{2}},...,h_{N(u)_{L}}))$
\STATE $h^{k}_{u}=\sigma(W^{k}\cdot CONCAT(h^{k-1}_{u},h^{k}_{N(u)}))$  
\ENDFOR
\STATE compute loss function $\mathcal{L}=\mathcal{L}_s+\mathcal{L}_r$
\STATE compute gradient of loss function $\nabla \mathcal{L}$ and update weight matrices $W$.
\ENDWHILE
\STATE \textbf{return} embedded representation $z_u=h^{K}_u/||h^{K}_{u}||,\forall{u}\in V$
\end{algorithmic}
\end{algorithm}

\subsection{System Architecture}
\label{architecture}
The system architecture is explicitly shown in Figure \ref{MF_gcn}. The input to our \textit{MF-GCN} aggregator is the center node attribute vector as well as all of its neighborhood nodes attributes. We choose the network layer depth $K=1$, since we find that one layer of Multi-Filtering GCN can already achieve good performance. We pick our local GCN filter number to be 25, and filter size to be 16. We concatenate these filters to form a 400 dimension aggregation representation, then we concatenate again with the center node attribute, and feed into a fully connected encoding layer to generate the embedding vector. We set the dimension off the embedding vector to be 100. This embedding vector is used as the final feature representation for nodes.

\section{Experiments}

\subsection{Experimental Setup}
To empirically evaluate the effectiveness and efficiency of our proposed model, we apply \textit{MF-GCN} in four public benchmark datasets: Citeseer, Cora, PubMed and Wiki. Citeseer, Cora and PubMed are three paper citation networks, where each node represents a paper and each edge represents a citation relationship. Wiki is a Wikipedia hyperlink network, where each node represents a web page and each edge indicates a hyperlink from one page to another. The node attributes of these networks are extracted as bag-of-words representation. Table \ref{dataset} presents the specific statistics of these four datasets. 
\begin{table}[b]
\centering
\begin{tabular}{lrrrrr}
\hline
 &Citeseer  & Cora & PubMed & Wiki \\
 \hline
\# Nodes      &3,312  &2,708 &19,717  &2,405   \\
\# Edges  &4,660 &5,278 &44,327 &17,981\\
Feature Dimension &3,703 &1,433 &500 &4,973\\
Node classes &6 &7 &3 &17\\
\hline
\end{tabular}
\caption{Statistics of datasets}
\label{dataset}
\end{table}

\textbf{Baselines} We compare our method against several well known baselines: DeepWalk, node2vec, LINE, role2vec, Raw feature (raw attributes are used as the embedding representation), GraphSAGE-GCN, GraphSAGE-mean, GraphSAGE-pool, and DANE (Deep Attribute Network Embedding). For sampling context nodes of a center node, we use the random walk strategy based on node2vec, and set $q=1$, $p=0.5$. For LINE, we adopt the second order proximity. Role2vec uses logarithmic binning and node attribute concatenation for role mapping functions. GraphSAGE-GCN is a special version of our \textit{MF-GCN} model where the filter number is 1. For GraphSAGE-Mean, and GraphSAGE-pool, we implement the algorithms according to the original paper, and we set the fixed GraphSAGE neighborhood sampling number to be 20. We set the latent representation dimension for all the comparing methods to be 100. We choose the architecture for our \textit{MF-GCN} model the same as described in section \ref{architecture}.
\subsection{Link Prediction}
Our first evaluation task is Link Prediction, which we test the models on all four benchmark datasets. Specifically, for each dataset, we randomly remove $50\%$ of the edges while keeping the graph connected, and train the model on the remaining graph. For training, since we do not reach node labels, we only use SGNN loss $\mathcal{L}_{s}$ as the cost function. For testing, we use all the removed edges as the positive testing edges, and we select equal number of nodes pairs from the network which have no edge connecting them as the negative testing edge; then we test the models on predicting the positive testing edge, as well as detecting the non-existed edges. The AUC scores of different methods are collected to reflect how well each model performs, shown in Table \ref{lp_auc}. From the experiment results, \textit{MF-GCN} generally performs better among all baseline methods.

\begin{table}[t]
\centering
\begin{tabular}{lcccc}
\hline
Model  &Citeceer &Cora &PubMed &Wiki \\
\hline
DeepWalk       & 0.771 &0.734 &0.857 &0.747   \\
Node2vec       & 0.721 &0.726 &0.868 &0.761    \\
LINE    &0.772 &0.723 &0.871 &0.728    \\
Role2vec    &0.921 &0.896 &0.898 &0.848    \\
\hline
GraphSAGE-mean  &0.917  &0.859  &0.936 & 0.869 \\
GraphSAGE-GCN &0.907 &0.892 &0.901 &0.880\\
GraphSAGE-pool &\textbf{0.937} &0.891 &0.932 &0.875\\
DANE &0.914 &0.844 &0.927 &0.821\\
\hline
\textit{MF-GCN} &0.924 &\textbf{0.913} & \textbf{0.944} &\textbf{0.893}\\
\hline
\end{tabular}
\caption{AUC score for link prediction on different dataset}
\label{lp_auc}
\end{table}

\subsection{Node Classification}
Our second evaluation task is node classification. We separately select $10\%$, $30\%$, $50\%$ nodes from the network as the 3 training sets, then test the classification accuracy on the remaining nodes. Micro-f1 score is used as the performance measurement. The testing results are shown in Table \ref{nc_citeceer}, \ref{nc_cora}, \ref{nc_pubmed}, and \ref{nc_wiki} respectively. From the test results, \textit{MF-GCN} shows significant improvement against all the baselines. Especially when the training data is limited, our \textit{MF-GCN} greatly outperforms others.
\begin{table}[h]
\centering
\begin{tabular}{lccc}
\hline
Model  & 10\% & 30\% &50\% \\
\hline
DeepWalk       & 0.368 &0.508  &  0.575   \\
Node2vec       & 0.373 &0.507 &0.604    \\
LINE    &0.389 &0.517 & 0.574    \\
Role2vec    &0.518 &0.659 & 0.699    \\
\hline
Raw Feature   & 0.551  &0.693  & 0.705    \\
GraphSAGE-mean  &0.627  &0.698  & 0.699 \\
GraphSAGE-GCN &0.647 &0.696 &0.703\\
GraphSAGE-pool &0.623 &0.706 &0.716\\
DANE &0.546 &0.703 &0.722\\
\hline
\textit{MF-GCN} &\textbf{0.686} & \textbf{0.713} &\textbf{0.746}\\
\hline
\end{tabular}
\caption{F1 score for node classification on Citeceer}
\label{nc_citeceer}
\end{table}

\begin{table}[h]
\centering
\begin{tabular}{lccc}
\hline
Model  & 10\% & 30\% &50\% \\
\hline
DeepWalk       & 0.317 &0.694 & 0.773    \\
Node2vec       & 0.338 &0.683 & 0.794     \\
LINE    &0.309 &0.697 &0.784     \\
Role2vec    &0.420 &0.635 &0.715     \\
\hline
Raw Feature   & 0.369  &0.652  & 0.734     \\
GraphSAGE-mean  &0.379  &0.793 & 0.796 \\
GraphSAGE-GCN &0.406 &0.772 &0.798\\
GraphSAGE-pool &0.439 &0.778 &0.788\\
DANE &0.502 &0.787 &0.804\\
\hline
\textit{MF-GCN} &\textbf{0.578} &\textbf{0.814} &\textbf{0.815}\\
\hline
\end{tabular}
\caption{F1 score for node classification on Cora}
\label{nc_cora}
\end{table}

\begin{table}[h]
\centering
\begin{tabular}{lccc}
\hline
Model  & 10\% & 30\% &50\% \\
\hline
DeepWalk       & 0.582 &0.709 & 0.725    \\
Node2vec       & 0.593 &0.705 & 0.727     \\
LINE    &0.566 &0.719 &0.731     \\
Role2vec    &0.616 &0.716 &0.733     \\
\hline
Raw Feature   & 0.637  &0.709  & 0.724     \\
GraphSAGE-mean  &0.739  &0.768 & 0.778 \\
GraphSAGE-GCN &0.679 &0.742 &0.773\\
GraphSAGE-pool &0.671 &0.749 &0.764\\
DANE &0.711 &0.758 &0.801\\
\hline
\textit{MF-GCN} &\textbf{0.758} &\textbf{0.771} &\textbf{0.813}\\
\hline
\end{tabular}
\caption{F1 score for node classification on PubMed}
\label{nc_pubmed}
\end{table}

\begin{table}[!h]
\centering
\begin{tabular}{lccc}
\hline
Model  & 10\% & 30\% &50\% \\
\hline
DeepWalk       & 0.148 &0.428 & 0.506    \\
Node2vec       & 0.156 &0.413 & 0.527    \\
LINE    &0.158 &0.424 &0.522     \\
Role2vec    &0.397 &0.562 &0.625     \\
\hline
Raw Feature   & 0.211  &0.423  & 0.532     \\
GraphSAGE-mean  &0.414  &0.466 & 0.568 \\
GraphSAGE-GCN &0.425 &0.552 &0.601\\
GraphSAGE-pool &0.432 &0.558 &0.560\\
DANE &0.434 &0.567 &0.604\\
\hline
\textit{MF-GCN} &\textbf{0.466} &\textbf{0.597} &\textbf{0.644}\\
\hline
\end{tabular}
\caption{F1 score for node classification on Wiki}
\label{nc_wiki}
\end{table}

\subsection{Optimized Filter Size}
\begin{figure}[t]
        \begin{center}
            \includegraphics[scale=0.25]{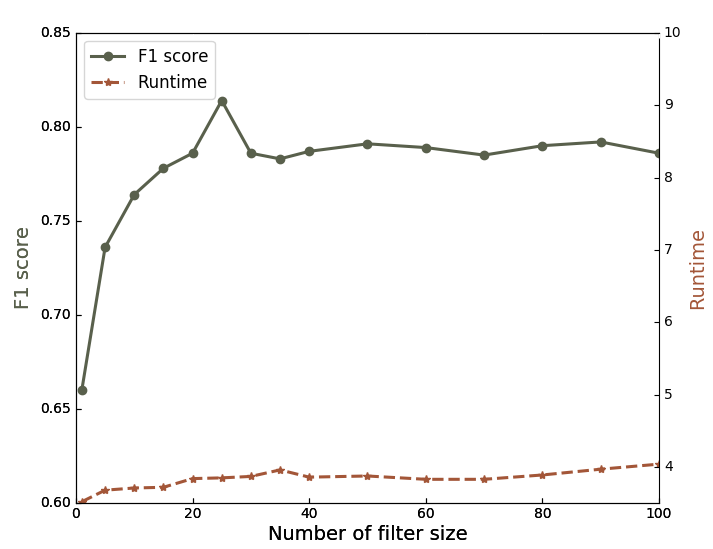}
        \end{center}
        \caption{Evaluation of effectiveness on different filter number}
        \label{filter_num}
\end{figure}

In order to show the effect of using different number of local GCN filters on node classification task, we conduct experiments of testing the F1-score on $70\%$ of test nodes from Cora under various number of filters (from filter number 1 to filter number 100) for training. We also collect different training time for one iteration under different number of filters. Our training is performed on a work station with two RTX 2080 Ti GPU, and Intel Core i9-9820X. Figure \ref{filter_num} shows the testing result. We find that the F1 score constantly increases from $0.66$ to $0.81$ with the filter number changes from 1 to 25; then it drops to 0.78, and remains relatively the same score as the filter number keeps increasing. We think the reason for this change is that adding more filters when the filter size is small could help capture different aspect of node features, thus leads to the increase of F1 score; but when the filter number reaches a saturation point, the system starts to be overfitting as the filter number keeps rising. Therefore, we pick the filter number to be 25 for our \textit{MF-GCN} model to achieve the best performance.
\begin{figure}[t]
        \begin{center}
            \includegraphics[scale=0.55]{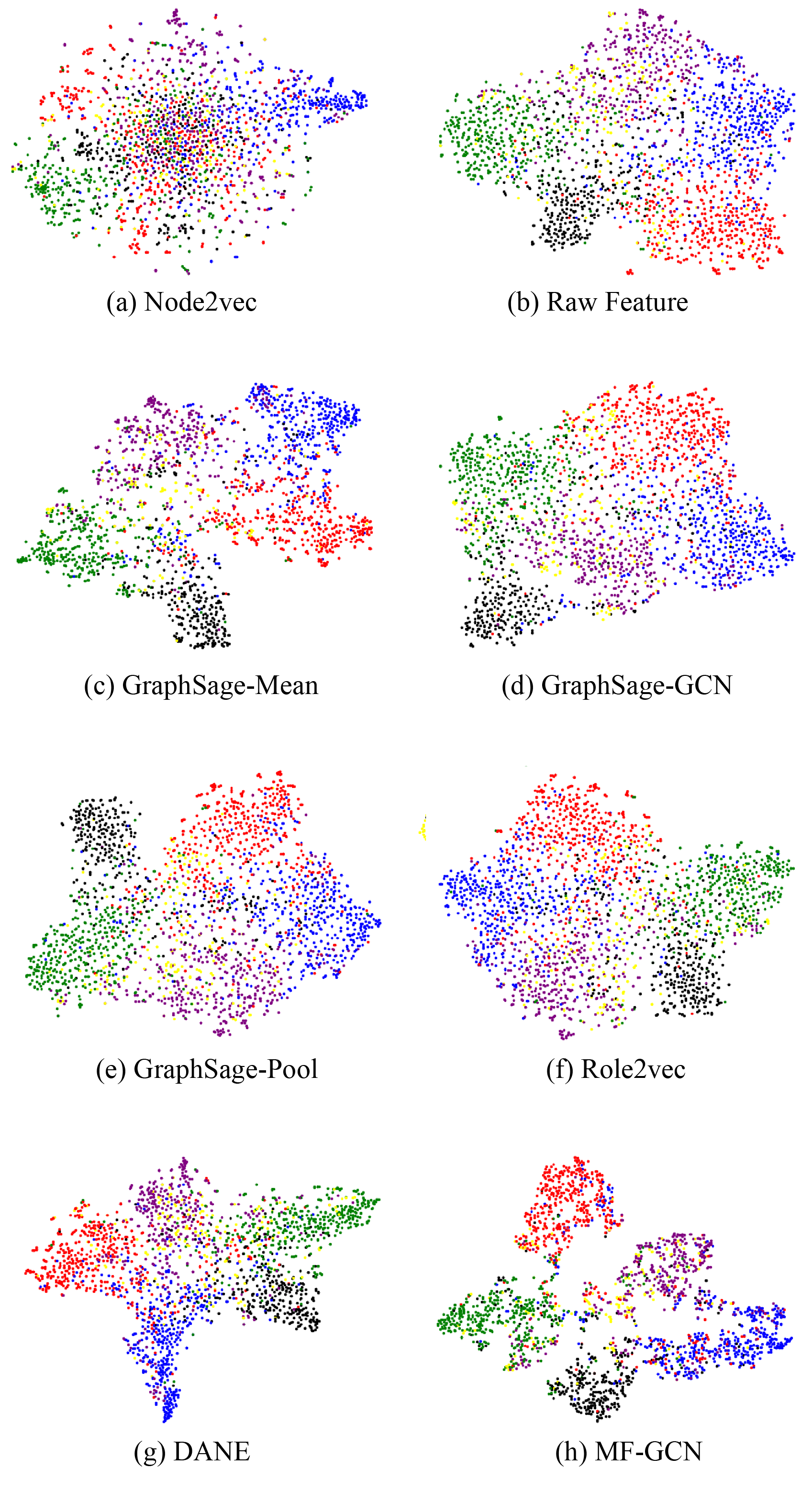}
        \end{center}
        \caption{2 dimensional embedding visualizations on citeceer}
        \label{2D_Embedding_Visualization_citeceer}
\end{figure}
\begin{figure}[t]
        \begin{center}
            \includegraphics[scale=0.55]{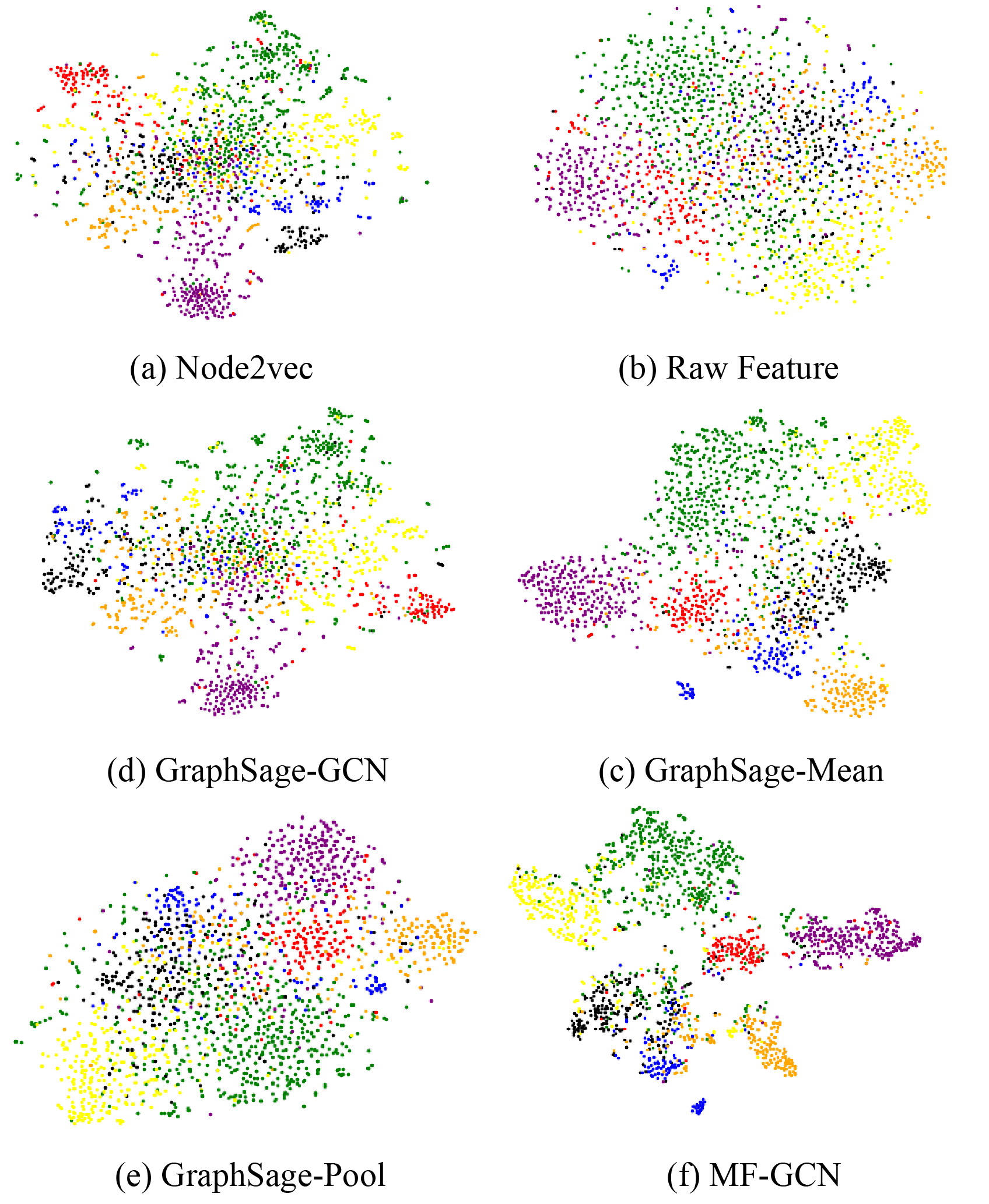}
        \end{center}
        \caption{2 dimensional embedding visualizations on Cora}
        \label{2D visualization}
\end{figure}
\begin{figure}[t]
        \begin{center}
            \includegraphics[scale=0.55]{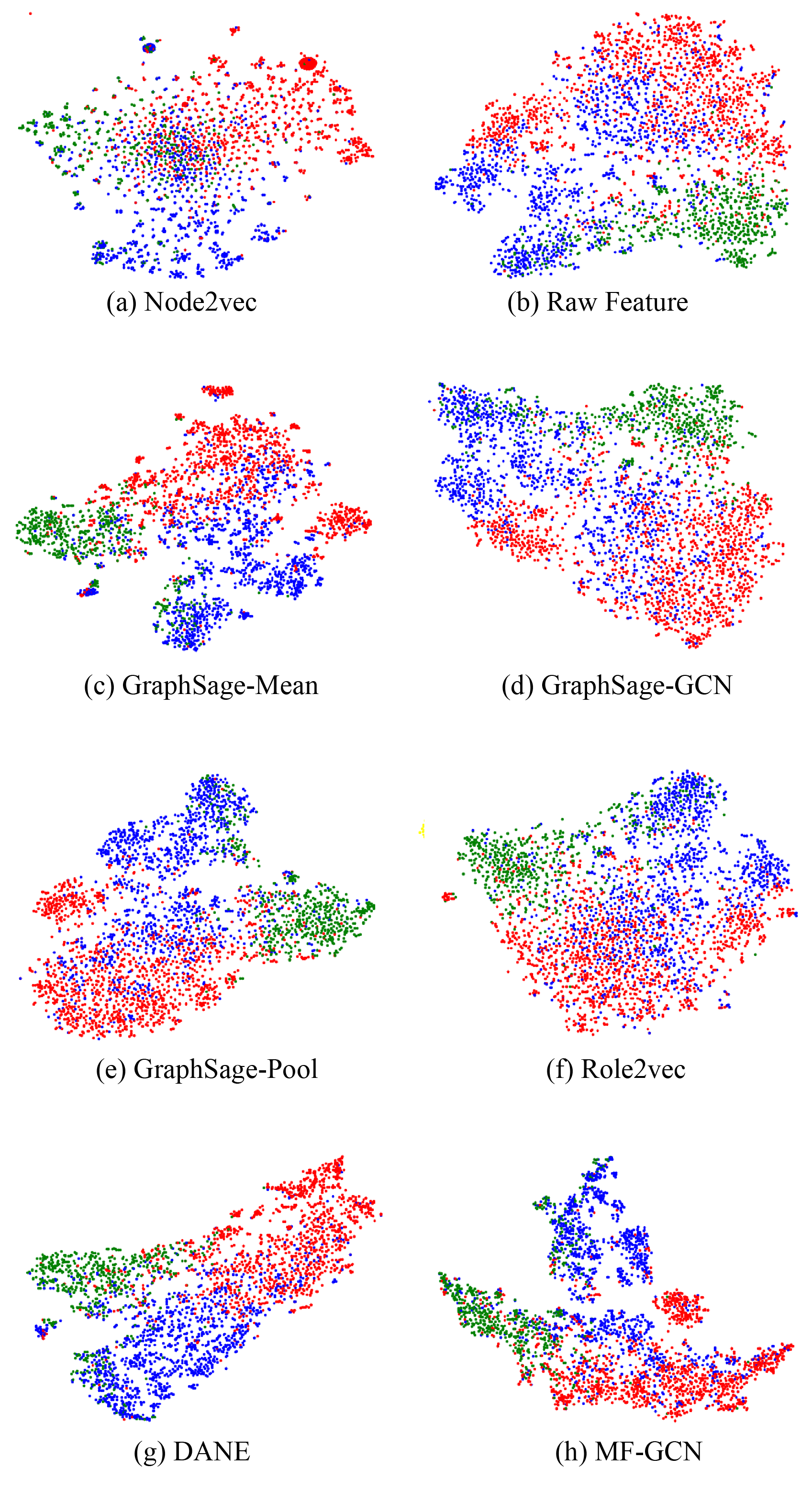}
        \end{center}
        \caption{2 dimensional embedding visualizations on PubMed}
        \label{2D_Embedding_Visualization_pubmed}
\end{figure}
\begin{figure}[t]
        \begin{center}
            \includegraphics[scale=0.55]{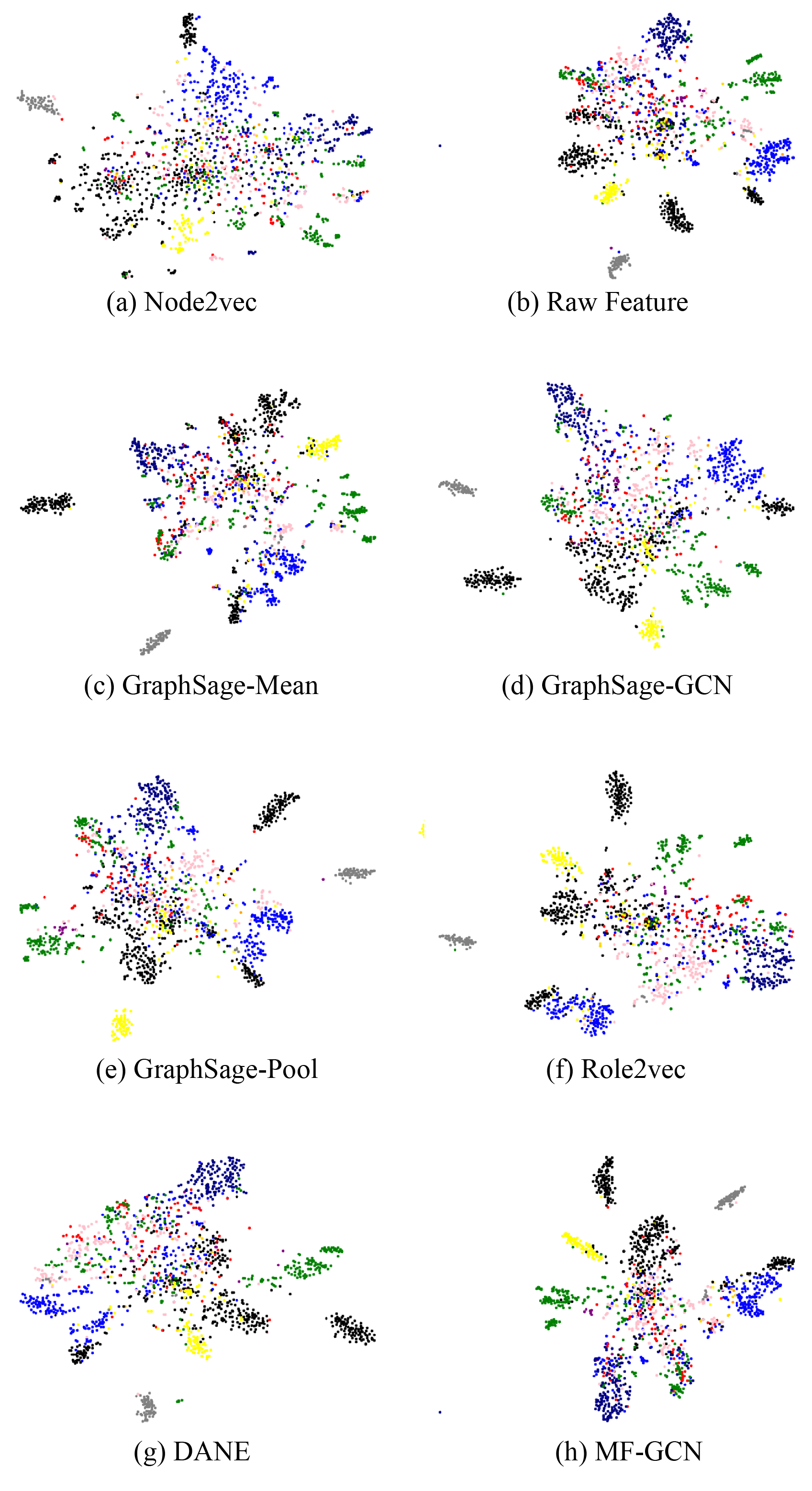}
        \end{center}
        \caption{2 dimensional embedding visualizations on Wiki}
        \label{2D_Embedding_Visualization_wiki}
\end{figure}

\subsection{2D visualization}
To visually present the advantage of our \textit{MF-GCN} model, we plot the 2 Dimensional embedding visualization of different methods using t-SNE \cite{maaten2008visualizing} as shown in Figure \ref{2D_Embedding_Visualization_citeceer}, \ref{2D visualization}, \ref{2D_Embedding_Visualization_pubmed} amd \ref{2D_Embedding_Visualization_wiki}. These plots are generated by performing node embedding on the $70\%$ test nodes from Cora, where we use $30\%$ nodes for training. From the plot, \textit{MF-GCN} clearly shows better performance on distinguishing different clusters of nodes against other baselines.

\section{Conclusion}
In this paper, we propose \textit{MF-GCN}, a novel network embedding approach that extracts different aspects of node features by using multiple local GCN filters. To show the effectiveness of our model, we conduct various experiments. First, we show \textit{MF-GCN} could increase performance of AUC score on link prediction task against many baseline methods. Second, we conduct experiments on node classification task from four public benchmark datasets and show that by using \textit{MF-GCN} model, the F1 score has a significant improvement against baseline methods, especially when the dataset is limited.

To show the effectiveness of using different numbers of filters, we conduct experiments to measure the performance of using different number of filters on node classification task, and provide suggestions on choosing the optimal number of filters.

Finally, the embedding results of \textit{MF-GCN} against other baseline methods are visually shown, which demonstrates that our \textit{MF-GCN} model has superior performance over other baseline methods.

For future directions, we would like to continue our research on incorporating attention mechanism that selectively picks important filters. 

\newpage
\appendix

\bibliographystyle{named}
\bibliography{ijcai20}

\begin{thebibliography}{}

\bibitem[\protect\citeauthoryear{Chen \bgroup \em et al.\egroup
  }{2018a}]{chen2018harp}
Haochen Chen, Bryan Perozzi, Yifan Hu, and Steven Skiena.
\newblock Harp: Hierarchical representation learning for networks.
\newblock In {\em Thirty-Second AAAI Conference on Artificial Intelligence},
  2018.

\bibitem[\protect\citeauthoryear{Chen \bgroup \em et al.\egroup
  }{2018b}]{chen2018fastgcn}
Jie Chen, Tengfei Ma, and Cao Xiao.
\newblock Fastgcn: fast learning with graph convolutional networks via
  importance sampling.
\newblock {\em arXiv preprint arXiv:1801.10247}, 2018.

\bibitem[\protect\citeauthoryear{Defferrard \bgroup \em et al.\egroup
  }{2016}]{defferrard2016convolutional}
Micha{\"e}l Defferrard, Xavier Bresson, and Pierre Vandergheynst.
\newblock Convolutional neural networks on graphs with fast localized spectral
  filtering.
\newblock In {\em Advances in neural information processing systems}, pages
  3844--3852, 2016.

\bibitem[\protect\citeauthoryear{Dong \bgroup \em et al.\egroup
  }{2017}]{dong2017metapath2vec}
Yuxiao Dong, Nitesh~V Chawla, and Ananthram Swami.
\newblock metapath2vec: Scalable representation learning for heterogeneous
  networks.
\newblock In {\em Proceedings of the 23rd ACM SIGKDD international conference
  on knowledge discovery and data mining}, pages 135--144. ACM, 2017.

\bibitem[\protect\citeauthoryear{Grover and
  Leskovec}{2016}]{grover2016node2vec}
Aditya Grover and Jure Leskovec.
\newblock node2vec: Scalable feature learning for networks.
\newblock In {\em Proceedings of the 22nd ACM SIGKDD international conference
  on Knowledge discovery and data mining}, pages 855--864. ACM, 2016.

\bibitem[\protect\citeauthoryear{Hamilton \bgroup \em et al.\egroup
  }{2017a}]{hamilton2017inductive}
Will Hamilton, Zhitao Ying, and Jure Leskovec.
\newblock Inductive representation learning on large graphs.
\newblock In {\em Advances in Neural Information Processing Systems}, pages
  1024--1034, 2017.

\bibitem[\protect\citeauthoryear{Hamilton \bgroup \em et al.\egroup
  }{2017b}]{hamilton2017representation}
William~L Hamilton, Rex Ying, and Jure Leskovec.
\newblock Representation learning on graphs: Methods and applications.
\newblock {\em IEEE Data(base) Engineering Bulletin}, 40:52--74, 2017.

\bibitem[\protect\citeauthoryear{Hammond \bgroup \em et al.\egroup
  }{2011}]{hammond2011wavelets}
David~K Hammond, Pierre Vandergheynst, and R{\'e}mi Gribonval.
\newblock Wavelets on graphs via spectral graph theory.
\newblock {\em Applied and Computational Harmonic Analysis}, 30(2):129--150,
  2011.

\bibitem[\protect\citeauthoryear{Huang \bgroup \em et al.\egroup
  }{2017}]{huang2017accelerated}
Xiao Huang, Jundong Li, and Xia Hu.
\newblock Accelerated attributed network embedding.
\newblock In {\em Proceedings of the 2017 SIAM international conference on data
  mining}, pages 633--641. SIAM, 2017.

\bibitem[\protect\citeauthoryear{Kipf and Welling}{2016}]{kipf2016semi}
Thomas~N Kipf and Max Welling.
\newblock Semi-supervised classification with graph convolutional networks.
\newblock {\em arXiv preprint arXiv:1609.02907}, 2016.

\bibitem[\protect\citeauthoryear{Levy and Goldberg}{2014}]{levy2014neural}
Omer Levy and Yoav Goldberg.
\newblock Neural word embedding as implicit matrix factorization.
\newblock In {\em Advances in neural information processing systems}, pages
  2177--2185, 2014.

\bibitem[\protect\citeauthoryear{Li \bgroup \em et al.\egroup
  }{2018}]{li2018adaptive}
Ruoyu Li, Sheng Wang, Feiyun Zhu, and Junzhou Huang.
\newblock Adaptive graph convolutional neural networks.
\newblock In {\em Thirty-Second AAAI Conference on Artificial Intelligence},
  2018.

\bibitem[\protect\citeauthoryear{Lin \bgroup \em et al.\egroup
  }{2015}]{lin2015learning}
Yankai Lin, Zhiyuan Liu, Maosong Sun, Yang Liu, and Xuan Zhu.
\newblock Learning entity and relation embeddings for knowledge graph
  completion.
\newblock In {\em Twenty-ninth AAAI conference on artificial intelligence},
  2015.

\bibitem[\protect\citeauthoryear{Maaten and
  Hinton}{2008}]{maaten2008visualizing}
Laurens van~der Maaten and Geoffrey Hinton.
\newblock Visualizing data using t-sne.
\newblock {\em Journal of machine learning research}, 9(Nov):2579--2605, 2008.

\bibitem[\protect\citeauthoryear{Mikolov \bgroup \em et al.\egroup
  }{2013}]{mikolov2013efficient}
Tomas Mikolov, Kai Chen, Greg Corrado, and Jeffrey Dean.
\newblock Efficient estimation of word representations in vector space.
\newblock {\em arXiv preprint arXiv:1301.3781}, 2013.

\bibitem[\protect\citeauthoryear{Niepert \bgroup \em et al.\egroup
  }{2016}]{niepert2016learning}
Mathias Niepert, Mohamed Ahmed, and Konstantin Kutzkov.
\newblock Learning convolutional neural networks for graphs.
\newblock In {\em International conference on machine learning}, pages
  2014--2023, 2016.

\bibitem[\protect\citeauthoryear{Ou \bgroup \em et al.\egroup
  }{2016}]{ou2016asymmetric}
Mingdong Ou, Peng Cui, Jian Pei, Ziwei Zhang, and Wenwu Zhu.
\newblock Asymmetric transitivity preserving graph embedding.
\newblock In {\em Proceedings of the 22nd ACM SIGKDD international conference
  on Knowledge discovery and data mining}, pages 1105--1114. ACM, 2016.

\bibitem[\protect\citeauthoryear{Perozzi \bgroup \em et al.\egroup
  }{2014}]{perozzi2014deepwalk}
Bryan Perozzi, Rami Al-Rfou, and Steven Skiena.
\newblock Deepwalk: Online learning of social representations.
\newblock In {\em Proceedings of the 20th ACM SIGKDD international conference
  on Knowledge discovery and data mining}, pages 701--710. ACM, 2014.

\bibitem[\protect\citeauthoryear{Qiu \bgroup \em et al.\egroup
  }{2019}]{qiu2019netsmf}
Jiezhong Qiu, Yuxiao Dong, Hao Ma, Jian Li, Chi Wang, Kuansan Wang, and Jie
  Tang.
\newblock Netsmf: Large-scale network embedding as sparse matrix factorization.
\newblock In {\em The World Wide Web Conference}, pages 1509--1520. ACM, 2019.

\bibitem[\protect\citeauthoryear{Scarselli \bgroup \em et al.\egroup
  }{2008}]{scarselli2008graph}
Franco Scarselli, Marco Gori, Ah~Chung Tsoi, Markus Hagenbuchner, and Gabriele
  Monfardini.
\newblock The graph neural network model.
\newblock {\em IEEE Transactions on Neural Networks}, 20(1):61--80, 2008.

\bibitem[\protect\citeauthoryear{Tang \bgroup \em et al.\egroup
  }{2015}]{tang2015line}
Jian Tang, Meng Qu, Mingzhe Wang, Ming Zhang, Jun Yan, and Qiaozhu Mei.
\newblock Line: Large-scale information network embedding.
\newblock In {\em Proceedings of the 24th international conference on world
  wide web}, pages 1067--1077. International World Wide Web Conferences
  Steering Committee, 2015.

\bibitem[\protect\citeauthoryear{Veli{\v{c}}kovi{\'c} \bgroup \em et al.\egroup
  }{2017}]{velivckovic2017graph}
Petar Veli{\v{c}}kovi{\'c}, Guillem Cucurull, Arantxa Casanova, Adriana Romero,
  Pietro Lio, and Yoshua Bengio.
\newblock Graph attention networks.
\newblock {\em arXiv preprint arXiv:1710.10903}, 2017.

\bibitem[\protect\citeauthoryear{Yang \bgroup \em et al.\egroup
  }{2016}]{yang2016revisiting}
Zhilin Yang, William~W Cohen, and Ruslan Salakhutdinov.
\newblock Revisiting semi-supervised learning with graph embeddings.
\newblock {\em arXiv preprint arXiv:1603.08861}, 2016.

\bibitem[\protect\citeauthoryear{Zhang \bgroup \em et al.\egroup
  }{2018}]{zhang2018anrl}
Zhen Zhang, Hongxia Yang, Jiajun Bu, Sheng Zhou, Pinggang Yu, Jianwei Zhang,
  Martin Ester, and Can Wang.
\newblock Anrl: Attributed network representation learning via deep neural
  networks.
\newblock In {\em IJCAI}, volume~18, pages 3155--3161, 2018.

\end{thebibliography}

\end{document}